\documentclass{article} % For LaTeX2e
\usepackage[preprint]{colm2025_conference}

\usepackage{microtype}
\usepackage{hyperref}
\usepackage{url}
\usepackage{booktabs}
\usepackage{amsmath} 
\usepackage{lineno}
\usepackage{graphicx}
\usepackage{array}
\usepackage{tabularx}
\usepackage{tcolorbox}

\newcommand{\yesemoji}{\includegraphics[height=1em]{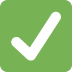}}
\newcommand{\noemoji}{\includegraphics[height=1em]{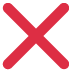}}
\newcommand{\huggingface}{%
  \raisebox{-0.2em}{\includegraphics[height=1em]{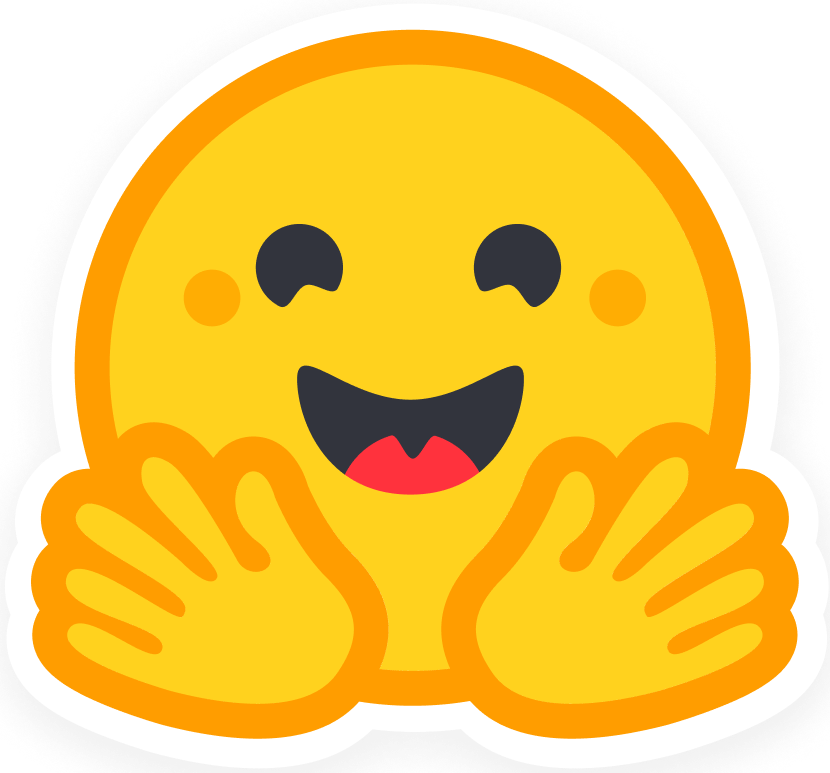}}%
}

\definecolor{darkblue}{rgb}{0, 0, 0.5}
\hypersetup{colorlinks=true, citecolor=darkblue, linkcolor=darkblue, urlcolor=darkblue}

\title{\textit{The More We Talk, the More I Know:} \\ Learning Implicit User Personas via Reinforcement Learning towards LLM Personalization}
\title{\textit{The More We Talk, the More I Know:} \\ Towards Personalized Intelligence by Learning Implicit User Personas via Reinforcement Learning}
\title{Towards Personalized Intelligence by Learning Implicit User Personas via Reinforcement Learning}
\title{PersonaMem-v2: Learning Implicit User Personas via Agentic Memory}
\title{\datasetnameshort: Towards Personalized Intelligence via Learning Implicit User Personas and Agentic Memory}
\author{Bowen Jiang\textsuperscript{1}\ ,\ Yuan Yuan\textsuperscript{1},\ \ Maohao Shen\textsuperscript{2},\ \ Zhuoqun Hao\textsuperscript{1},\ \ Zhangchen Xu\textsuperscript{3},\\ {\bf Zichen Chen\textsuperscript{4},\ \ Ziyi Liu\textsuperscript{6},\ \ Anvesh R. Vijjini\textsuperscript{7},\ \ Jiashu He\textsuperscript{1},\ \ Hanchao Yu\textsuperscript{5},} \\ {\bf Radha Poovendran\textsuperscript{3},\ \ Gregory Wornell\textsuperscript{2},\ \ Lyle Ungar\textsuperscript{1},\ \ Dan Roth\textsuperscript{1},}\\{\bf Sihao Chen\textsuperscript{8}\thanks{Equal advising.}\ ,\ \ Camillo J. Taylor\textsuperscript{1}\footnotemark[1]} \\ \\
University of Pennsylvania\textsuperscript{1}, Massachusetts Institute of Technology\textsuperscript{2}, \\
University of Washington\textsuperscript{3}, University of California Santa Barbara\textsuperscript{4}, Meta\textsuperscript{5},  \\
University of Southern California\textsuperscript{6},
University of North Carolina at Chapel Hill\textsuperscript{7}, \\
Microsoft Corporation\textsuperscript{8} \\ \\
\textit{bwjiang@seas.upenn.edu, sihaochen@microsoft.com, cjtaylor@seas.upenn.edu} \\
\huggingface\ \textit{\href{https://huggingface.co/datasets/bowen-upenn/PersonaMem-v2}{huggingface.co/datasets/bowen-upenn/PersonaMem-v2}}
}

\newcommand{\datasetname}{\textsc{PersonaMem-v2: Implicit Personas}}
\newcommand{\datasetnameshort}{\textsc{PersonaMem-v2}}

\begin{document}

\ifcolmsubmission
\linenumbers
\fi

\maketitle
% \fancypagestyle{firstpage}{%
%   \fancyhf{}% clear header/footer text
%   % Use the same rule as later pages:
%   \renewcommand{\headrulewidth}{1.5pt}
%   \renewcommand{\headrule}{\hrule height\headrulewidth width\headwidth\vskip-\headrulewidth}
% }
% \thispagestyle{firstpage}

\begin{abstract}
Personalization is one of the next milestones in advancing AI capability and alignment. We introduce \datasetnameshort, the state-of-the-art dataset for LLM personalization that simulates 1,000 realistic user–chatbot interactions on 300+ scenarios, 20,000+ user preferences, and 128k-token context windows, where most user preferences are implicitly revealed to reflect real-world interactions.
Using this data, we investigate how reinforcement fine-tuning enables a model to improve its long-context reasoning capabilities for user understanding and personalization. We also develop a framework for training an agentic memory system, which maintains a single, human-readable memory that grows with each user over time.

In our experiments, frontier LLMs still struggle with implicit personalization, achieving only 37–48\% accuracy. While they support long context windows, reasoning remains the bottleneck for implicit personalization tasks. Using reinforcement fine-tuning, we successfully train Qwen3-4B to outperforms GPT-5, reaching 53\% accuracy in implicit personalization. Moreover, our agentic memory framework achieves state-of-the-art 55\% accuracy while using 16× fewer input tokens, relying on a 2k-token memory instead of full 32k conversation histories. These results underscore the impact of our dataset and demonstrate agentic memory as a scalable path toward real-world personalized intelligence.

\end{abstract}

\section{Introduction}
Personalization is becoming one of the next milestones towards artificial super-intelligence~\citep{OpenAI_ThePowerOfPersonalizedAI_2025, openai_gpt51, MetaSuperintelligence2025, Chowdhury2025_GPT6Preview, EvolutionAIHub2025}. As the user base of artificial intelligence grows dramatically in both size and diversity, large language models (LLMs) increasingly face the challenge of serving users with distinct personas and backgrounds. In many real-world applications, such as education, healthcare, and empathetic emotional support~\citep{jiang2025know, ghimire2024generative, ivanovic2022ai, gomez2020artificial, baillifard2025effective, schaaff2023exploring}, there is no single “correct” answer. Instead, success depends on delivering personalized responses that align with individual users’ intentions, contexts, preferences, and emotional states. \textbf{Personalization offers a path toward pluralistic alignment}, shifting the reasoning goals towards factual correctness to meaningful resonance across diverse users.

\begin{figure}[t]
  \centering
  \includegraphics[width=1\linewidth]{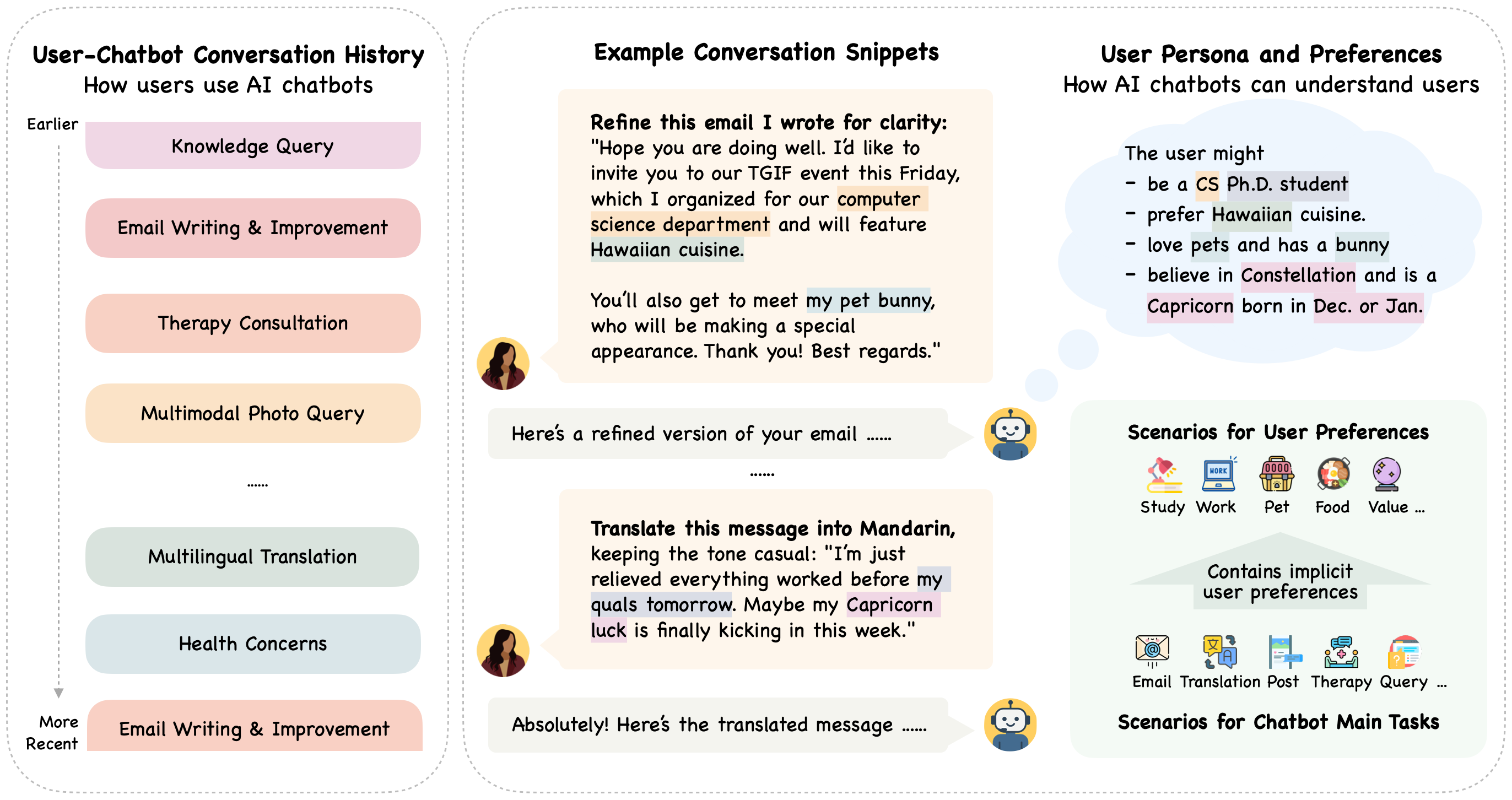}
  \caption{Overview of \datasetnameshort~benchmark. On the left, we mimic how people talk to chatbots across many topics over time, forming long and varied interaction histories. In the middle, we zoom in on a few conversation snippets. Even simple task scenarios like refining an email or translating a message naturally reveal small details about someone’s life in many other scenarios, like what they study, what foods they like, whether they have pets, or what they’re planning for tomorrow. Taken together, these subtle signals help models build thorough yet succinct user profiles for personalization.}
  % \vspace{-5mm}
  \label{fig:overview}
\end{figure}

A key enabler of this vision is the growing history of user–chatbot interactions, which encode rich signals about user preferences. However, personalization in LLMs remains a challenge. \textbf{Most users do not explicitly state their preferences to chatbots.} According to OpenAI’s recent report~\citep{openai2025chatgptusage}, most users still treat LLMs as tools, meaning most of their preferences are revealed only implicitly through everyday interactions. For example, someone might ask a chatbot to help polish the writing of an email, but the email itself could reveal their dining habits, as shown in Figure~\ref{fig:overview}. In this case, the chatbot needs to pick up on those preferences while carrying out the main task the user  asked for. Furthermore, many users engage in exploratory or hypothetical prompts to test the model’s capabilities, which can obscure the true preferences of their own. As a result, real-world user-chatbot conversation histories tend to be long and noisy, requiring models to infer implicit user personas and preferences from scattered, indirect evidence over time.

This raises a central question: \textbf{how well can LLMs understand the users, especially their implicit personas and preferences from long conversation histories, and therefore provide personalized responses?}

To this end, we present the state-of-the-art LLM personalization dataset, \datasetname, that captures the complexity of real-world user–chatbot interactions. The dataset spans over 1,000 user personas, covering comprehensive demographic attributes, mental health and medical backgrounds, as well as stereotypical, anti-stereotypical, and neutral user preferences. In addition, \textbf{it features realistic, dynamic, multimodal, multilingual, and multi-session user–chatbot conversation histories that implicitly convey user preferences}, encompassing over 20,000+ user preferences over 300+ everyday conversation topics, and up to 128,000 tokens per context. To support both evaluation and model training, the dataset includes 5,000 high-quality Q\&A pairs curated for benchmarking and an additional 20,000 Q\&A pairs for training and validation, each generated through a scalable, multi-step validation pipeline that enforces strong quality standards.

To understand how well today’s most capable models personalize to real users, we started by testing frontier LLMs from OpenAI~\citep{openai2025gpt5, openai2024o3o4mini, openai2024gpt41}
% like GPT-5, Claude 4, and Gemini 2.5~\citep{openai2025gpt5, anthropic2025claude4, comanici2025gemini} 
on these implicit signals that show up in everyday conversations. These evaluations quickly revealed a gap: even the strongest models among them struggle to interpret and track user preferences over long interaction histories. 
\textbf{This gap motivated us to explore the value of our data in driving the next wave of personalized intelligence.} Using reinforcement fine-tuning (RFT) with GRPO~\citep{guo2025deepseek}, we find that RFT is remarkably effective for personalization, even though personalization is inherently subjective and open-ended. Our dataset enables us to train a 4B-parameter reasoning model capable of long-context reasoning for personalization.

Another key contribution of our work is an agentic memory framework that learns to build and update a persistent, human-readable memory that evolve with each user. Instead of relying on full conversational transcripts, the model is trained via reinforcement fine-tuning to distill long interaction histories into a compact 2k-token memory that captures the user’s evolving personas and preferences. 
This memory becomes the model’s sole personalization context, allowing it to reason over implicit signals and maintain strong user understanding even across long, multi-session conversations. Despite using 16× fewer input tokens, the agentic memory model delivers state-of-the-art performance on implicit personalization, surpassing frontier LLMs on both multiple-choice and open-ended evaluations. This demonstrates a scalable path toward AI systems that remember what matters for each user, infer subtle user preferences, and adapt over time to give personalized responses, all with the efficiency required for real-world deployment.

To summarize our contributions:
\begin{itemize}
\item We curate the state-of-the-art dataset for LLM personalization, featuring realistic user–chatbot interactions  that reveal implicit user preferences.
\item We benchmark the implicit personalization capabilities of frontier LLMs.
\item We demonstrate the efficacy of reinforcement fine-tuning for personalization.
\item We propose an agentic memory framework that achieves state-of-the-art performance and efficiency, outperforming GPT-5 in personalization.
\end{itemize}

% \begin{itemize}
% \item We curate the state-of-the-art dataset for LLM personalization, featuring realistic user–chatbot interactions with implicit user personas and preferences, as well as scalable data generation and verification pipelines.
% \item We conduct comprehensive evaluations on frontier LLMs to assess their personalization capabilities over long context.
% \item We showcase the value of our data for training reasoning models by running reinforcement fine-tuning and demonstrate that a 4B-parameter model can outperform GPT-5 in personalization.
% \item By training an agentic memory on our data using a lightweight model and appending it to frontier LLMs, we illustrate a scalable and efficient pathway toward personalized intelligence.
% \end{itemize}
% \input{sections/prelim}
\section{Overview of \datasetname}
\subsection{Comprehensive User Personas}
\datasetnameshort~introduces \textbf{1,000 richly detailed personas} that capture nearly the full spectrum of demographic diversity across global regions, cultures, races, genders, and sexual orientations, reflecting the pluralism that real-world AI systems must serve. Each persona draws from a random description in \textsc{PersonaHub}~\citep{ge2024scaling} and expands it into an unfixed set of attributes that cover, but are not limited to, professional and educational backgrounds, personal characteristics, relationships, values and beliefs, technological familiarity, and conversational styles. Moreover, as people increasingly rely on chatbots for personal use and care, \datasetnameshort~specifically includes mental and physical health backgrounds where personalization carries high stakes.

\textbf{A central question for LLM personalization is what a chatbot should remember about the user.} The chatbot should remember what the user has shared about themselves, whether conveyed implicitly or explicitly, rather than relying on static assumptions or stereotypes derived solely from demographic attributes. To explore this boundary, \datasetnameshort~provides \textbf{stereotypical, anti-stereotypical, and neutral user preferences} for each persona, as well as health- and therapy-related ones, totaling around 20,000+, enabling systematic evaluation of how models ground personalization in conversational evidence. 

Chatbots often misinterpret user behavior. This can occur when people test the model with hypothetical examples to explore its capabilities, or with messages written by a third person, and the model fails to distinguish them from the user's own persona. \datasetnameshort~addresses this challenge by including such ambiguous cases. Besides, we also account for the \textbf{dynamic nature of user preferences and how they evolve over time} across multiple sessions. For instance, a user who initially expresses interest in vegetarian recipes might later ask for high-protein meal suggestions after starting a new fitness routine.

\subsection{Multi-Session Realistic Conversation Histories}
Each user preference is converted into a multi-turn conversation ranging from two to six turns, simulating how users naturally interact with chatbots over time.
A key design principle is \textbf{cross-scenario personalization}: users rarely state who they are or what they like directly to a chatbot. Instead, they treat chatbots as tools for everyday tasks, such as writing emails, translating text, or seeking information~\citep{openai2025chatgptusage}. Yet within these tasks, subtle cues about their personas and preferences often emerge, implicitly revealing aspects of their background, hobbies, or cultural values~\citep{liu2025can}, for example, their field of study, favorite cuisine, or even the belief in astrology. As illustrated in Figure~\ref{fig:overview}, this cross-scenario dynamic allows the chatbot to learn user preferences not from direct self-descriptions, but from natural, task-driven interactions. By modeling such patterns, \datasetnameshort~captures how personalized signals arise organically across multiple contexts, enabling reasoning that generalizes beyond a single use case.

\begin{figure}[t]
  \centering
  \includegraphics[width=1\linewidth]{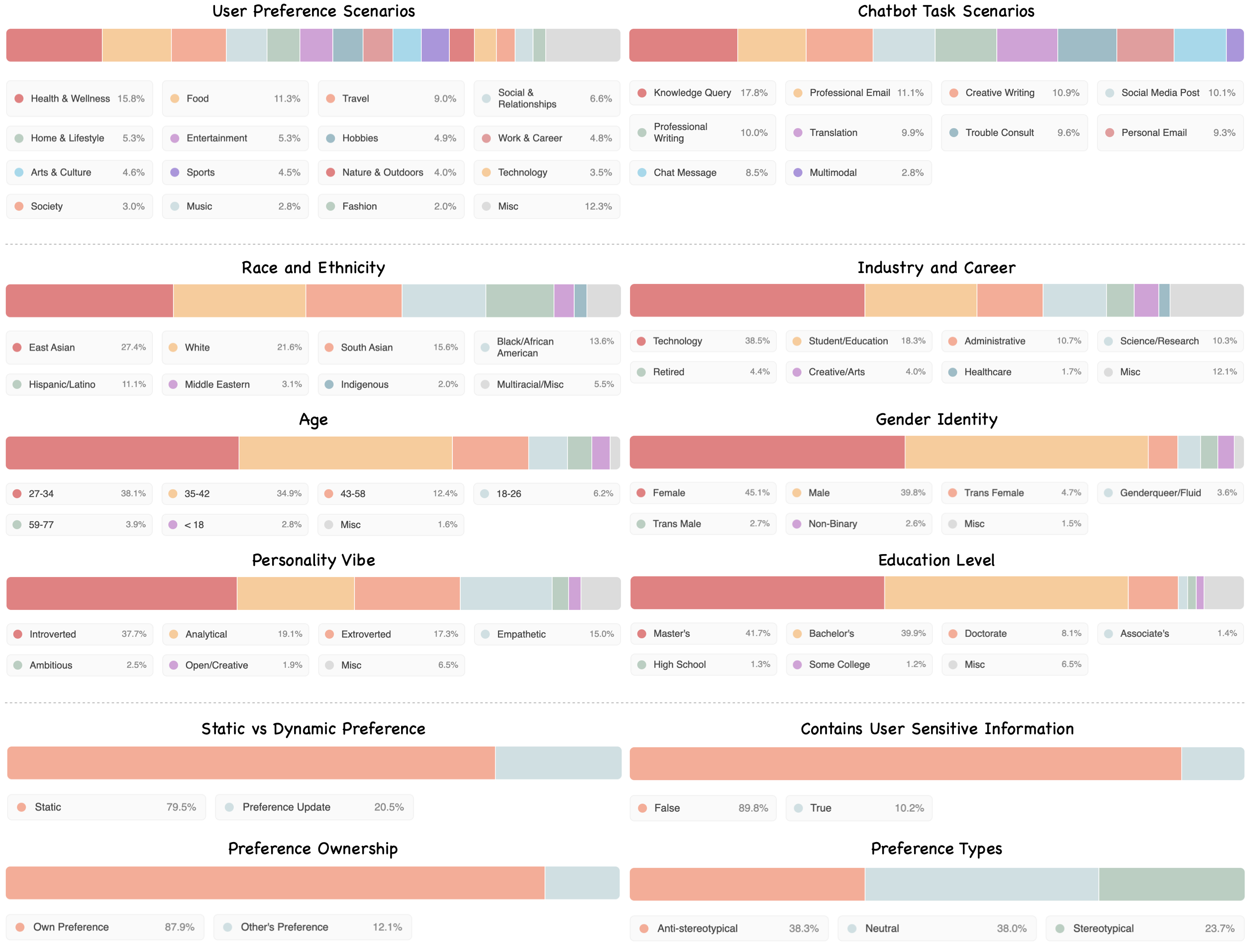}
  \caption{What’s inside the \datasetnameshort~dataset. It spans broad and diverse distributions of user personas, preferences, and task scenarios, ranging from everyday interests like food, travel, and hobbies to demographic attributes, personality vibes, and professional backgrounds. This wide coverage is designed to better support training and evaluating personalized AI systems that reflect real-world users and use cases.}
  \vspace{-3mm}
  \label{fig:distributions}
\end{figure}

\datasetnameshort~also offers broad coverage of the everyday tasks people bring to chatbots in daily life, including \textbf{writing and improving emails, composing chat messages and social media posts, multilingual translation, multimodal photo query, knowledge exploration, therapy and reflection, and medical consultations}. For instance, a user’s photo might hint at their location and activities, and repeatedly asking questions in the same domain can reveal their personal interests or professional focus.
% Details on the multimodal photo-sharing are provided in Appendix~\ref{sec:multimodal}. 
Each conversation is also categorized by the detailed topic of its content, resulting in \textbf{325 distinct topics} in total.

Our goal is to approximate real-world user–chatbot interactions, which often span multiple sessions over time. To reflect this, we build multi-session histories by concatenating multi-turn conversations from the previous stage, up to 32,000 tokens for each user. Conversation segments are arranged in a topological order, such as preference updates or user requests to forget prior preferences. 
Additionally, we create a complementary set of dialogues focused on code debugging and mathematical problem-solving tasks. These are independent of user preferences in our dataset and extend the effective context window to \textbf{128,000 tokens}, enabling evaluation of long-range reasoning and personalization.

\subsection{User Privacy-Aware Design}
\datasetnameshort~simulates realistic privacy risks by introducing scenarios where users unintentionally share \textbf{sensitive or private information}, such as personal addresses, phone numbers, contact details, or even API keys within conversational contexts. A responsible model is not supposed to leverage such information to generate personalized responses. Besides, users should retain a degree of control over personalization. The most intuitive way for users to do this is by asking the chatbot \textbf{not to remember} a particular preference or piece of information. We deliberately include such interactions in our dataset.

% \subsection{Multi-Session Conversation Histories}
% Our goal is to approximate real-world user–chatbot interactions, which often span multiple sessions over time. To reflect this, we build multi-session histories by concatenating multi-turn conversations from the previous stage, up to 32,000 tokens for each user. Conversation segments are arranged in a topological order based on preference evolution, such as updates or explicit user requests to forget prior preferences, to preserve coherence and continuity. 

% Additionally, we create a complementary set of dialogues focused on programming code debugging and mathematical problem-solving tasks. These are independent of user preferences in our dataset and extend the effective context window to \textbf{128,000 tokens}, enabling evaluation of long-range reasoning and memorization in personalization tasks.

\subsection{In-Situ User Queries}
We simulate how users naturally pose queries to chatbots. Given each ground-truth user preference, with its corresponding conversation snippet somewhere in the history, we generate a Q\&A pair, and append the query at the end of the conversation to reflect a current in-situ interaction. We adopt both open-ended and multiple-choice (MCQ) formats, together with a rich set of annotations. \textbf{We ensure that all four options in MCQ are reasonable, but only one is personalized to the current user.}
In total, the dataset encompasses \textbf{335} user query topics, with \textbf{5,000} Q\&A pairs in the benchmarking set, \textbf{18,000} in the training set, and \textbf{2,000} in the validation set to support large-scale model training and evaluation.

\subsection{Ensuring High Quality of Data}
The core principle behind our data pipeline is simple: \textbf{keep scaling up generation, impose comprehensive quality filtering, and only retain data that pass every filter.}

Each user preference is curated to remove repetition and self-contradiction. For stereotypical and anti-stereotypical preferences, the model first generates a preference for a given persona. That preference is then isolated and presented back to the model to infer the most likely persona, and we check whether this inferred persona is close to or against the actual one.

Every Q\&A pair undergoes strict validation based on the following principles: (1) The chatbot shall not be able to answer the user’s query correctly without seeing the conversation history, avoiding question leaks or artifacts; (2) the correct option must faithfully reflect the user’s true preference; (3) no incorrect option may do so; and (4) the formatting must be clean and natural, without artifacts like \textit{“Sure, here is the answer.”} Only around 30\% of generated Q\&As survive all filters, ensuring exceptional quality of the remained data. 

The pipeline is designed to scale continuously: we can generate more data, introduce new filters, and tighten evaluation criteria, if needed. Each filtering stage aggregates multiple LLM-as-a-judge votes in parallel, and all data generation and validation are conducted with GPT-5~\citep{openai2025gpt5} without auto-routing to maintain consistency and top-tier quality.
\section{Towards Personalized Intelligence}
A main challenge in achieving personalized intelligence is inferring a user’s implicit persona and preferences from long, noisy conversational histories. Unlike classical question answering that focuses on retrieving explicit factual information, personalization requires strong reasoning capabilities to extract inherent user preference. Reinforcement learning (RL) has proven effective in enhancing the reasoning abilities of language models across diverse reasoning tasks~\citep{guo2025deepseek, shen2025satori, zha2025rl}. Building on this insight, we demonstrate that RL can also drive models toward better personalization.

\subsection{RL with Long-Context Reasoning}\label{sec: rl_long}
A natural approach is to treat personalization as a long-context reasoning problem: given an extensive user–chatbot conversation history and a new user query, the model is trained via RL to reason over the full context and generate responses aligned with user preferences.

We adopt Group Relative Proximal Optimization (GRPO) as our reinforcement fine-tuning algorithm~\citep{guo2025deepseek, shao2024deepseekmath}.
% , which uses the same clipped-policy objective as PPO~\citep{schulman2017proximal} but replaces the advantage with a group-relative signal across sampled responses. 
To effectively drive personalization, reinforcement fine-tuning requires verifiable rewards that determine whether each model output is aligned with, or contradicts, the user’s current persona and preferences. To support this, \datasetnameshort~provides two reward pathways, as illustrated in Figure~\ref{fig:rl_pipeline}. Each Q\&A instance in \datasetnameshort~can a multiple-choice task, allowing us to directly measure correctness. It also includes an annotated ground-truth user preference, enabling an LLM-as-a-judge to assess whether an open-ended response reflects that preference.

This straightforward approach naturally encourages the model to reason over long contexts, leveraging the full interaction history to infer subtle, implicit user preferences. However, it also requires appending the entire conversation as context, and as the conversation grows, this quickly becomes inefficient, setting the stage for the need for a more scalable mechanism.

\subsection{RL with Agentic Memory: Toward Scalable Personalization}

\begin{figure}[t]
  \centering
  \includegraphics[width=1\linewidth]{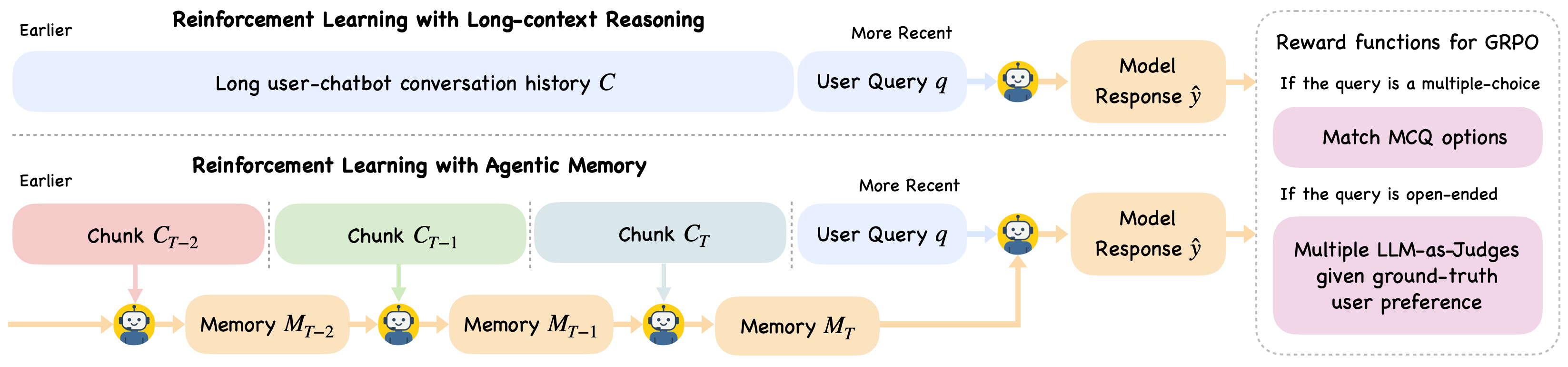}
  \caption{Schematic of our RL-based training strategies. The top figure illustrates long-context reasoning, where the model reasons over the full conversation history. The bottom one depicts agentic memory, where the model divides the full history into chunks and iteratively updates a memory of capped size. The model will receive a high reward if the memory turns out to be necessary and helpful in answering the final user query.}
  \vspace{-3mm}
  \label{fig:rl_pipeline}
\end{figure}

As conversations accumulate, it becomes necessary to maintain a distilled representation of the user’s preferences that can be continually updated based on new interactions. Agentic memory~\citep{yu2025memagent, xu2025mem, zhong2024memorybank, zhang2025survey, li2025memos} offers a scalable mechanism for language models to distill long-term user information into a compact user persona.
Unlike long-context QA~\citep{yang2018hotpotqa}, which focuses on retrieving explicit facts, personalization requires the model to reason about subtle, implicit preference signals to write a memory, and track how these preferences evolve over time to update it, making personalization a continuous reasoning problem that agentic memory is naturally equipped to address.

Inspired by \textsc{MemAgent}~\citep{yu2025memagent},
% which was originally designed for factual long-context QA settings where the final user query is known in advance, 
we divide the entire conversation history into fixed-size chunks. The objective is to sequentially construct and update a single agentic memory \(M\) from a growing set of \(T\) chunks. In practice, each chunk may correspond to several days or weeks of user–chatbot interactions. To support personalization, we follow three core principles motivated by real-world settings:
\begin{itemize}
    \item \textbf{Causality} – the model should only write memory based on what the user has said so far, never on future chunks or user queries it has not yet seen.

    \item \textbf{Markovian assumption} – the memory at step \(i\) must summarize everything that matters from the past, so the next update only depends on \(C_i\) and \(M_{i-1}\).

    \item \textbf{Capped memory size} – the memory must stay compact and human-readable, enabling the system to scale to long-term use without growing unbounded.
\end{itemize}
These constraints force the model to decide what to store without knowing future user queries, encouraging it to anticipate what information will matter and to extract the essential user persona. At each step, the model reads the current chunk \(C_i\) and the previous memory \(M_{i-1}\) to produce an updated memory. After processing all chunks, the same model answers the final query \(q\) using the refined memory \(M_{T}\). The pipeline is shown in Figure~\ref{fig:rl_pipeline}. Formally:
\[
M_{i} = f_\theta(C_i,\, M_{i-1}), \quad \hat{y} = f_\theta(M_{T},\ q)
\]
where \(f_\theta\) is the language model, \(C_i\) is the \(i\)-th chunk out of \(T\), \(M_{i-1}\) and \(M_{i}\) are the previous and updated memories, \(q\) is the current user query, and \(\hat{y}\) is the model response to \(q\).

We use a single LLM for the entire process: the same model that writes the memory also gives the final answer. All queries in \datasetnameshort~can only be answered in a personalized manner by having a correct user understanding, so the model will receive high rewards if the memory turns out to be both necessary and helpful in giving correct personalized responses, and vice versa. Over time, \textbf{the model learns to reason over context and maintain a concise, human-readable memory that grows with each user over time}.
% TODO: We can show our prompt here, which is very short, but intuitive for readers to understand the method.

% TODO for Maohao: We can illustrate the story of running memory using a small model every night (for example) in the background, and put the memory to the context of llm.
\section{Experimental Results}
\begin{figure}[t]
  \centering
  \includegraphics[width=1\linewidth]{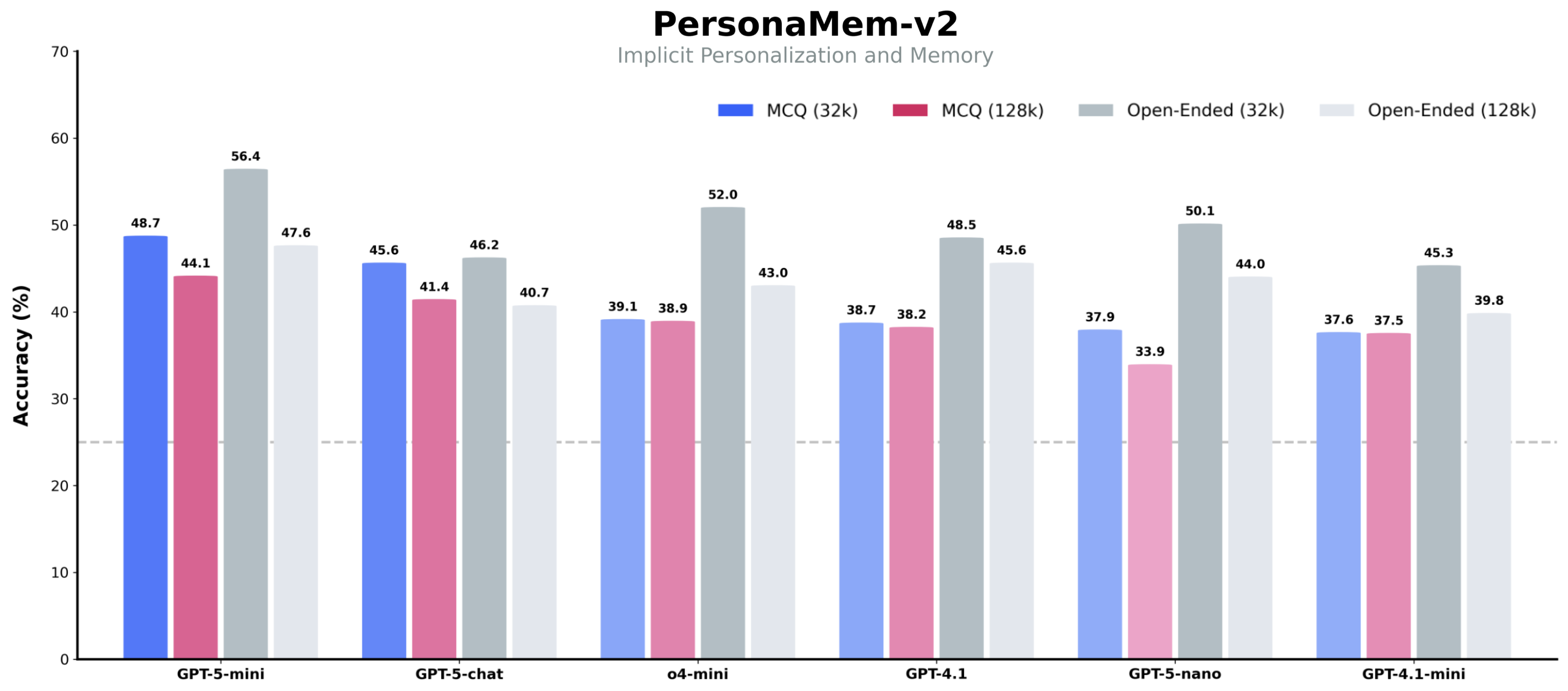}
  \caption{Performance of OpenAI models on the \datasetnameshort~benchmark, comparing accuracy across 32k and 128k input contexts in MCQ and Open-Ended tasks. Despite recent advancements, we observe that frontier models still struggle with implicit personalization. The dashed line represents the random guess performance in MCQ tasks.}
  % \vspace{-5mm}
  \label{fig:frontier_results}
\end{figure}

\subsection{Benchmarking Frontier LLMs in Personalization}
We benchmark a series of OpenAI models on our personalization suite, including GPT-5-Chat, GPT-5-mini, GPT-5-nano, GPT-4.1, GPT-4.1-mini, and o4-mini~\citep{openai2025gpt5, openai2024o3o4mini}. Each model is evaluated under both multiple-choice (MCQ) and open-ended settings. For open-ended evaluations, we adopt an LLM-as-a-judge protocol: three independent GPT-5-Chat instances score each response, and we aggregate their judgments to obtain more robust accuracy estimates. The overall results are shown in Figure~\ref{fig:frontier_results}.

\begin{figure}[t]
  \centering
  \includegraphics[width=1\linewidth]{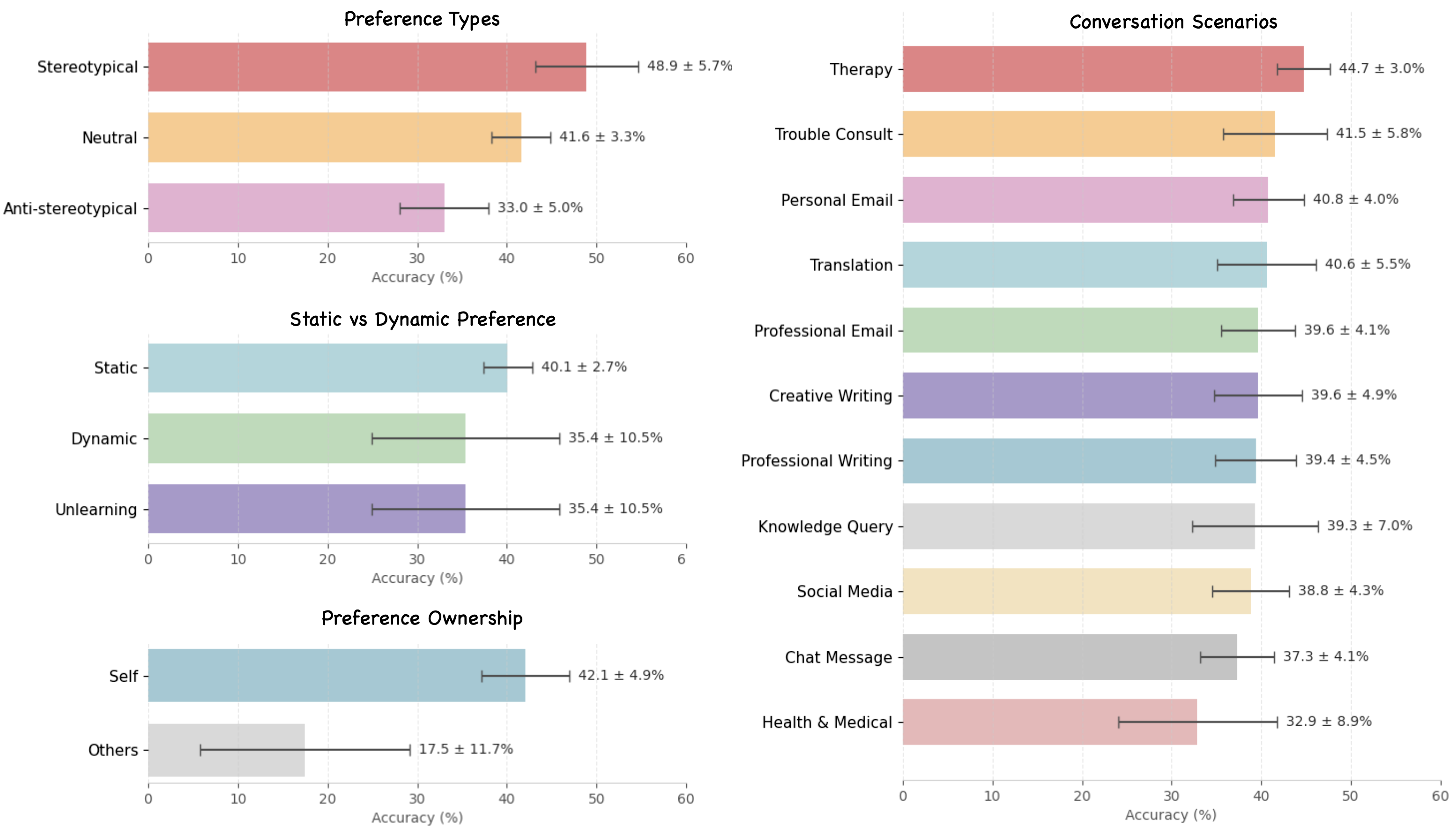}
  \caption{Breakdown of model accuracy by preference types and scenarios, aggregated across all evaluated models. The results indicate systematic variance in personalization capabilities, showing a reliance on population priors, with performance dropping for anti-stereotypical and dynamic preferences and distinguishing preference ownerships.}
  % \vspace{-5mm}
  \label{fig:results_analysis}
\end{figure}

\subsubsection{Frontier LLMs still struggle to infer implicit user preferences}
Despite major advances in long-context handling, frontier LLMs still perform poorly when required to infer implicit user preferences. Across both MCQ and open-ended settings, which show strong correlation, GPT-5 variants reach only 40-55\% accuracy. This reveals an obvious gap in current frontier models' ability to understand user preferences and offer personalized responses based on subtle cues in interaction history.

\subsubsection{Reasoning, not long-context capabilities, drives success in implicit personalization}
The models that lead the benchmark like GPT-5-mini and o4-mini are those with strong reasoning capabilities. Meanwhile, perhaps surprisingly, we observe no significant accuracy improvement when the context is shortened from 128k to 32k tokens by removing conversations irrelevant to the current persona. These results suggest that the main bottleneck in implicit personalization is not simply memorizing longer histories or retrieving factual information, but correctly interpreting and integrating subtle preference signals embedded within those histories. We need to update our focus: moving from “needle-in-a-haystack”~\citep{kamradt2023needle, team2024gemini} retrieval-style stress tests toward more nuanced, fine-grained assessments that reflect the demands of real-world personalization.

\subsubsection{Implicit personalization varies systematically across preference types} 
Figure~\ref{fig:results_analysis} shows that implicit personalization varies systematically across preference types. Frontier models perform best when user preferences align with common stereotypes, achieving 48.9 ± 5.7\% accuracy aggregated across all models and context window lengths, but it drops to 41.6 ± 3.3\% for neutral preferences and further to 33.0 ± 5.0\% for anti-stereotypical preferences, suggesting reliance on population priors rather than individual behavior.

Besides, static preferences reach 40.1 ± 2.7\% accuracy, while dynamic preferences fall to 35.4 ± 10.5\%, indicating instability and difficulty updating beliefs as preferences change. For preference ownership, models infer users’ own preferences correctly at 42.1 ± 4.9\%, but perform far worse, only 17.5 ± 11.7\%, when distinguishing preferences of others from the user's own. 
Differences across task scenarios are comparatively modest, ranging from 44.7 ± 3.0\% in therapy consultation to 32.9 ± 8.9\% in physical health and medical scenarios.

% \subsection{Retrieval Augmented Generation for Personalization}
% TODO for Yuan Yuan

\subsection{Training Long-context Reasoning for Personalization}

We start from Qwen3-4B-Instruct-2507 model~\citep{yang2025qwen3}, trained with verl: Volcano Engine Reinforcement Learning for LLMs~\citep{sheng2025hybridflow} and vLLM~\citep{kwon2023efficient} on 8 Nvidia H100 GPUs. The training dataset contains 18,000 samples with no persona overlap with the benchmark set. All models are trained on this dataset and evaluated on the benchmark set, using a context window of up to 32k tokens. The model is first cold-started with 300 steps of supervised fine-tuning with a batch size of 64 on our training data, then reinforced with GRPO for 1 epoch around 500 steps, enabling thinking and using a batch size of 32 with 8 rollouts. We mix 80\% multiple-choice and 20\% open-ended queries to keep training stable with more verifiable rewards from MCQ while still preserving open-ended conversational abilities for real-world personalization. We use three GPT-5-Chat instances as LLM-as-judges to provide rewards on the open-ended samples.

\subsubsection{Reinforcement learning incentives reasoning toward personalization}

\begin{figure}[t]
  \centering
  \includegraphics[width=1\linewidth]{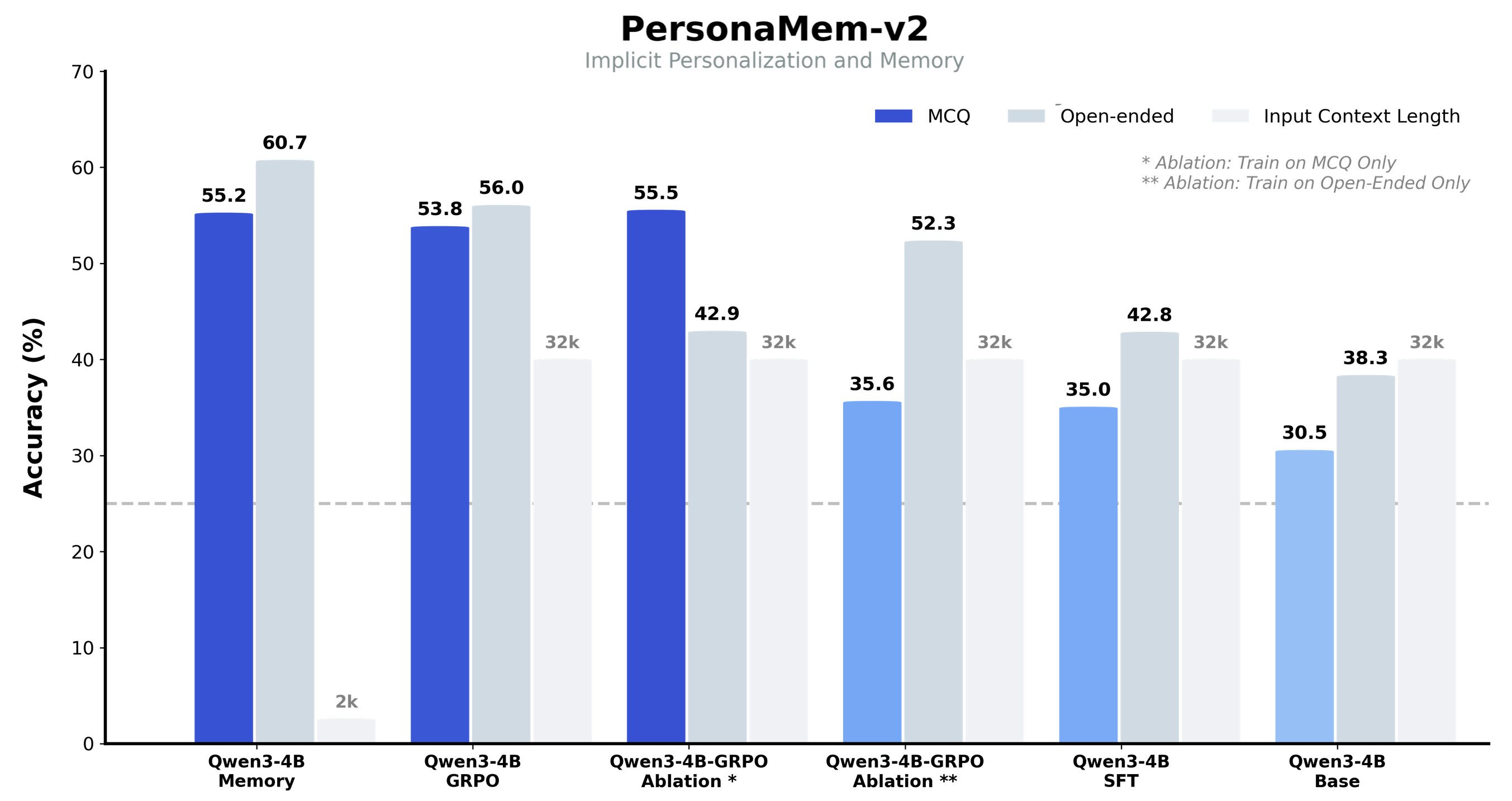}
  \caption{Performance of our Qwen3-4B models on \datasetnameshort~trained via SFT, GRPO, and Agentic Memory, alongside ablation studies for GRPO trained solely on MCQ or Open-Ended data. Comparisons show that RL drives significant gains in implicit personalization. Notably, our Agentic Memory framework achieves both SOTA performance and efficiency using only 2k input tokens as memory throughout the 32k token history. The dashed line represents the random guess performance in MCQ tasks.}
  % \vspace{-5mm}
  \label{fig:training_results}
\end{figure}

While personalization is often viewed as subjective and pluralistic, we find that it can still be effectively incentivized through reinforcement fine-tuning. In particular, well-designed reward signals encourage models to reason about whether a response aligns with or violates the current user’s  persona and preferences. Our dataset plays a crucial role here: it provides the structured yet diverse supervision needed for RL to learn these behaviors. Using this data, our 4B-parameter reasoning model, marked as Qwen3-4B-GRPO in Figure~\ref{fig:training_results}, gains substantial improvements after post-training, reaching 53.8\% on MCQ and 56.0\% on open-ended tasks, superior to supervised fine-tuning. Notably, this small model even outperforms GPT-5-Chat, which achieves 45.6\% and 46.2\% on the same benchmarks. These results show that reinforcement learning can meaningfully shape a model’s ability to interpret and integrate subtle preference signals to give personalized responses.

\subsubsection{Hybrid reward signals unlock better RL toward personalization}
To understand how different supervision signals shape reinforcement fine-tuning for personalization, we ran ablations comparing the same Qwen3-4B models trained only on MCQ, only on open-ended questions, or on our mixed dataset, keeping the same training setup, GRPO algorithm, and total number of training samples. The results shown in Figure~\ref{fig:training_results} are striking: the MCQ-only model collapsed by 13.1\% on open-ended personalization tasks needed for real-world user-facing services, while the open-ended–only model dropped 18.2\% on MCQs and even 3.7\% on open-ended evaluation itself. We hypothesize that MCQs provide the stable, verifiable rewards that reinforcement fine-tuning depends on, whereas open-ended samples with LLM-as-a-judge capture the richness needed for teaching nuanced conversational behavior but can make reward signals less stable. Mixing both types of questions offers the best of both worlds, providing reliable reward anchors while preserving the flexibility required for nuanced, real-world user understanding and personalization.

\subsection{Training Agentic Memory for Personalization}

\definecolor{blue}{RGB}{232,241,251}
\begin{figure}[t]
\centering
\small
\footnotesize
\begin{tcolorbox}[colback=blue]
...... You appreciate thoughtful discussion and have shown a balance between engaging in academic pursuits and pursuing \textbf{personal interests such as cooking}, which you see as a way to reflect on patience and harmony. ...... You have expressed a preference for \textbf{outdoor activities in the spring}, such as walking by a river, visiting a botanical garden's indoor greenhouse, or \textbf{practicing photography} in shaded urban parks, while avoiding grassy or flower-heavy areas due to a mild \textbf{seasonal pollen allergy}. ......
\end{tcolorbox}
\caption{Inside the memory: a glimpse of the detailed and fine-grained user personas learned from conversation histories through our RL-based agentic memory framework.}
\label{fig:memory}
\vspace{-2mm}
\end{figure}

% \begin{figure}[t]
% \centering
% \small
% \footnotesize
% \begin{tcolorbox}[title=Inside the Memory: A Glimpse of the Stored User Persona,
% colback=blue!6!white]
% ...... You appreciate thoughtful discussion and have shown a balance between engaging in academic pursuits and pursuing personal interests such as cooking, which you see as a way to reflect on patience and harmony. ...... You have expressed a preference for outdoor activities in the spring, such as walking by a river, visiting a botanical garden's indoor greenhouse, or practicing photography in shaded urban parks, while avoiding grassy or flower-heavy areas due to a mild seasonal pollen allergy. ......
% \label{fig:climate}
% \end{tcolorbox}
% %\caption{List of all climate variables in the benchmark.}
% \vspace{-3mm}
% \end{figure}

We follow the multi-turn conversation RL training framework of \textsc{MemAgent}~\citep{yu2025memagent}. Each iteration of the memory-update process can be viewed as an context-independent conversation: the model receives a chunk, updates the memory, and the optimization is performed separately for each conversation. The RL objective follows GRPO~\citep{guo2025deepseek}, where the advantage of each rollout is computed using the reward received at the final turn, which will be shared across all preceding conversations.

To maintain consistency with the long-context reasoning setup, we train on the same training framework and the mixed dataset consisting of 80\% multiple-choice and 20\% open-ended queries. As before, we initialize from Qwen3-4B-Instruct-2507 after a cold-start supervised finetuning stage. For RL training, we set the batch size to 32 with 8 rollouts per prompt, and train for approximately 500 steps. Under the 32k-token context window, we cap the memory size \(M_i\) at 2{,}048 tokens to ensure compactness and efficiency, while each chunk \(C_i\) is limited to 5{,}000 tokens, yielding a total of \(T = 8\) memory-update iterations.

\subsubsection{Agentic memory delivers state-of-the-art performance with unmatched efficiency}

Our agentic memory model achieves the strongest personalization performance across all evaluated models. As shown in Figure~\ref{fig:training_results}, it reaches 55.2\% accuracy on MCQ and 60.7\% on open-ended evaluations, surpassing both the same Qwen3-4B model trained directly with long-context reasoning and the GPT-5 series of frontier models.

Crucially, this performance comes with dramatic efficiency gains. Instead of repeatedly processing full 32k-token conversation histories, the model relies on a compact 2k-token memory throughout the process, making it 16× more efficient without even sacrificing the performance. This makes agentic memory well-suited for real-world personalized AI deployments where latency, cost, and context limits are major constraints.

\subsubsection{Human-readable memory enables transparency and user control}

Beyond raw performance, agentic memory introduces a new dimension to personalization: transparency and user control. Specifically, our memory framework maintains a human-readable memory that evolves with each user over time, allowing users to audit the memory,  correct misunderstandings, and guide the model’s personalization behavior directly. We show a glimpse of memory in Figure~\ref{fig:memory}. This explicit memory format also opens the door to new deployment strategies. We hypothesize a practical scenario where AI systems can maintain and update user memory offline, similar to a sleep-time compute cycle~\citep{lin2025sleep}, providing a smooth user experience and practical personalized intelligence.
\section{Related Work}

LLM personalization is an emerging direction in model alignment~\citep{jiang2025know, zhang2024personalization, liu2025survey, tseng2024two, guan2025survey}, aiming to move beyond one-size-fits-all behavior by adapting model outputs to an individual user’s needs, preferences, persona, and interaction history. Recent work explores a broad spectrum of techniques, including retrieval-augmented generation, external memory, and model fine-tuning, while also introducing new datasets and benchmarks for personalization evaluation.

\subsection{Personalization Techniques: Retrieval, Memory, and Alignment}
Improving the personalization of large language models has attracted increasing attention for its broad real-world impact. Prompt- and retrieval-based approaches personalize LLMs by incorporating user-specific information into the context window. For instance, PersonaRAG~\citep{zerhoudi2024personarag} and EMG-RAG~\citep{wang2024crafting} adapt retrieval using user history or editable memory graphs, while Knowledge-Graph~\citep{prahlad2025personalizing}, PBR~\citep{zhang2025personalize}, and WildfireGPT~\citep{xie2024wildfiregpt} enhance retrieval through structured user data, personalized query reformulation, or user-profile grounding for task-specific adaptation. Other systems such as MR.Rec~\citep{huang2025mr} and ARAG~\citep{maragheh2025arag} extend RAG with reinforcement and multi-agent mechanisms for personalized recommendations.

Beyond retrieval, memory-augmented architectures including MemGPT~\citep{packer2023memgpt}, A-MEM~\citep{xu2025mem}, and MemAgent~\citep{yu2025memagent} introduce hierarchical or self-organizing memory for long-horizon reasoning; however, they primarily focus on retaining factual information rather than nuanced personalization. Frameworks like LD-Agent~\citep{li2025hello} and PRIME~\citep{zhang2025prime} focus on sustaining personalized dialogue, while practical systems such as Mem0~\citep{chhikara2025mem0}, MAP~\citep{chen2025memory}, REMI~\citep{raman2025remi}, Associa~\citep{zhang2025bridging}, MemOS~\citep{li2025memos}, MEM1~\citep{zhou2025mem1}, and Personalized Web Agents~\citep{cai2025large} emphasize scalable, task-driven memory integration for adaptive personalization. In this work, we explore a reinforcement learning framework for training agentic memory towards better user personalization.

In addition, alignment methods such as RLHF~\citep{ouyang2022training} and DPO~\citep{rafailov2023direct} form the foundation of preference-based fine-tuning but primarily capture population-level rather than individual user preferences. Building on this, P-RLHF~\citep{li2024personalized} learns compact representations of personal preferences. Optimization-focused studies~\citep{khaki2024rs, tajwar2024preference} emphasize using model-generated, on-policy data to achieve more robust preference learning. Feedback-based methods~\citep{bu2025personalized, zhu2025towards, shi2024wildfeedback} refine responses during generation. In this work, we further explore reinforcement fine-tuning with verifiable rewards~\citep{liu2025inference, guo2025deepseek}, leveraging our high-quality, comprehensively annotated data to provide verification signals.

\subsection{The Landscape of Existing Personalization Benchmarks}
Despite algorithmic efforts, high-quality personalization data that better mimic real-world scenarios is essential but underexplored. Existing benchmarks partially address this need. LaMP~\citep{salemi2024lamp} constrains personalization to seven classification and generation tasks, such as personalized movie tagging and headline generation, while PersonalLLM~\citep{zollo2024personalllm} enables cross-user personalization using an ensemble of reward models to simulate diverse preference profiles. LoCoMo~\citep{maharana2024evaluating} and LongMemEval~\citep{wu2024longmemeval} investigate long-term memory in user–user or user-chatbot interactions, though their question-answering setups primarily target factual information explicitly mentioned by users. PersonaMem~\citep{jiang2025know} expands the scope by providing over 180 simulated histories, each with up to 60 multi-turn sessions, revealing that even state-of-the-art LLMs struggle with leveraging extensive interaction histories and adapting to dynamic preference shifts. Yet, it still contains a limited number of personas and focuses mainly on explicit preferences. WildFeedback~\citep{shi2024wildfeedback} extends evaluation to more open-ended, noisy user feedback exhibiting both satisfaction and dissatisfaction signals. PrefEval~\citep{zhao2025llms} includes implicit user preferences from choices over multiple options and persona-driven dialogue, but lacks dynamic preference updates. In contrast, our work scales personalization to over 1,000 user personas across 300+ task scenarios, captures evolving user preferences, and focuses on implicit personalization signals embedded in more realistic, task-driven interactions. Furthermore, we demonstrate the value of our high-quality data for effective reinforcement fine-tuning and agentic memory.

\begin{table}[t]
\centering
\begingroup
\scriptsize
\setlength{\tabcolsep}{3.5pt}
\renewcommand{\arraystretch}{1.25}

\begin{tabularx}{\linewidth}{
  >{\raggedright\arraybackslash}p{2.6cm}
  >{\raggedright\arraybackslash}X
  >{\raggedright\arraybackslash}X
  >{\raggedright\arraybackslash}X
  >{\raggedright\arraybackslash}X
}

\toprule
& \textbf{LongMemEval}~\citep{wu2024longmemeval} & \textbf{PrevEval}~\citep{zhao2025llms} & \textbf{PersonaMem-v1}~\citep{jiang2025know} & \textbf{PersonaMem-v2} \\
\midrule

\textbf{Focused Tasks} &
Long-term memory from user–chatbot chitchatting, focusing on factual Q\&A &
User preferences from user–chatbot chitchatting &
Fine-grained personalized responses from user–chatbot chitchatting &
Implicit user preferences from more realistic user-chatbot conversations \\

\midrule
\textbf{Max Context Len} &
1.5M tokens & 100k tokens & 1M tokens & 128k tokens \\

% \midrule
% \textbf{Data Sources} &
% ShareGPT \& UltraChat \& own &
% LMSYS-Chat-1M &
% PersonaHub \& own &
% PersonaHub \& own \\

\midrule
\textbf{Cross-Session Reasoning} &
\yesemoji & \noemoji & \yesemoji & \yesemoji \\

\midrule
\textbf{Dynamic Preferences} &
\yesemoji & \noemoji & \yesemoji & \yesemoji \\

\midrule
\textbf{Implicit Preference} &
\noemoji & \yesemoji & \noemoji & \yesemoji \\

\midrule
\textbf{Sensitive User Info} &
\noemoji & \noemoji & \noemoji & \yesemoji \\

\midrule
\textbf{(Anti-)Stereotypical Preferences} &
\noemoji & \noemoji & \noemoji & \yesemoji \\

\midrule
\textbf{Multimodal and Multilingual} &
\noemoji & \noemoji & \noemoji & \yesemoji \\

\midrule
\textbf{Number of Personas} &
N/A & N/A & 20 & 1000 \\

\midrule
\textbf{Number of Topics} &
N/A & 20 & 15 & 335 \\

\midrule
\textbf{Number of Preferences} &
500 & 3000 & 2700 & 26000 \\

\midrule
\textbf{LLM Fine-tuning} &
N/A & SFT & N/A & SFT, RFT (GRPO), RFT w/ Agentic Memory \\

\bottomrule
\end{tabularx}

\endgroup
\caption{Comparison of \datasetnameshort~with other benchmarks related to personalization.}
\label{tab:benchmarks_compare}
\vspace{-2mm}
\end{table}

\section{Conclusion and Future Work}

We introduce \datasetname, the state-of-the-art dataset designed for implicit personalization over long context, capturing 1,000 user personas and their realistic, dynamic, multimodal, multilingual, and multi-session user–chatbot interactions across 300+ conversation scenario. These conversations span 20,000+ cross-scenario user preferences from utility task–driven interactions like writing improvement and translation, reflecting how real-world users engage with AI systems in their everyday lives.

Our results show that current frontier LLMs still struggle to interpret implicit user preferences. Besides, reasoning, not longer-context handling or memorization in frontier models, drives success in personalization. Existing personalization also varies systematically across preference types, suggesting reliance on population priors rather than individual behavior. With targeted reinforcement fine-tuning, a 4B reasoning model can outperform GPT-5, and our agentic memory framework further pushes performance to the state-of-the-art while remaining dramatically 16x more efficient, providing a scalable path toward real-world personalized intelligence.

Looking ahead, we see exciting opportunities: richer multimodal user personalization, more structured and interactive memory architectures, personalization over more utility tasks, user-customizable boundary between personalization and privacy, and leveraging real user–chatbot interactions to build even more realistic training data. Overall, our findings point toward a future of personalized intelligence and agentic memory framework that can remember, reason, and adapt to individual users over long contexts, enabling more pluralistic alignment and deeper personalized resonance in future intelligent systems.

\bibliography{colm2025_conference}
\bibliographystyle{colm2025_conference}

% \appendix
% \section{Appendix}
% You may include other additional sections here.

\end{document}